\documentclass[twoside,leqno,twocolumn]{article}

\usepackage{ltexpprt} 
\usepackage{amsmath}
\usepackage{amssymb}
\usepackage{graphicx}
\usepackage{color}
\usepackage{bm}
\usepackage{wrapfig}
\usepackage{subfigure}
\usepackage{booktabs}

\usepackage[font={small}]{caption}

\begin{document}

\title{\Large Improving confidence while predicting trends in temporal disease networks}

\author{Djordje Gligorijevic\thanks{Center for Data Anaytics and Biomedical Informatics, Temple University  \{gligorijevic, jelena.stojanovic, zoran.obradovic\}@temple.edu} \\
\and 
Jelena Stojanovic\footnotemark[1] \\
\and
Zoran Obradovic\footnotemark[1]}

\date{}

\maketitle


\begin{abstract} \small\baselineskip=9pt 

For highly sensitive real-world predictive analytic applications such as healthcare and medicine, having good prediction accuracy alone is often not enough. These kinds of applications require a decision making process which uses uncertainty estimation as input whenever possible. Quality of uncertainty estimation is a subject of over or under confident prediction, which is often not addressed in many models. In this paper we show several extensions to the Gaussian Conditional Random Fields model, which aim to provide higher quality uncertainty estimation. These extensions are applied to the temporal disease graph built from the State Inpatient Database (SID) of California, acquired from the HCUP. Our experiments demonstrate benefits of using graph information in modeling temporal disease properties as well as improvements in uncertainty estimation provided by given extensions of the Gaussian Conditional Random Fields method.
\end{abstract}

\section{Introduction}

The increased availability of Electronic Health Record (EHR) data has created suitable conditions for various studies that are aimed for improvement of healthcare quality level. Data mining based studies on EHR data have shown the potential for improvement of healthcare systems \cite{Dey2014, Zhou2013}. These methods aim at building stable frameworks for estimating different states in healthcare systems and providing significant knowledge to healthcare practitioners. Predictive modeling application to health outcomes related to medical data in terms of diseases, procedures, mortality, and other measures may have a huge impact on quality of treatment, improve detection of high risk groups of patients, or detect important effects not taken into consideration in prior medical treatments. These are some of the problems that can be addressed by inspecting and analyzing medical data.

However, some of the problems these studies encounter are data locality. For example such cases are applications where data records are specific for particular hospitals (e.g. only general hospitals) or group of patients (e.g. Medicare patients). Therefore, it is very difficult to create models that provide good generalization performance for chosen applications \cite{Stiglic2014}. 
An additional problem in the analysis is posed by the heterogeneity of the data due to different data sources, often having different quality. This data heterogeneity can potentially compromise quality of conclusions derived from the given analysis. 

For these reasons, we focus on analysis of the Healthcare Cost and Utilization Project (HCUP)  database \cite{hcup}, particularly the SID database from the hospitals in the state of California. This database contains more than 35 million inpatient discharge records over 9 years and it is not specific to a group of hospitals but contains data for entire state. 
It can be induced that the data can be observed as non--local for California and any analysis can be generalized for that state. 
For each patient in the database there is demographic information (like age, birth year, sex, race), diagnosis (primary and up to 25 others), procedures (up to 25), information about hospital stays and other information (like length of stay, total charges, type of payment and payer, discharge month, survival information). Using this data could potentially give us insight in many healthcare problems.

Among many difficult tasks, capturing global trends of diseases are the ones that intrigue many healthcare practitioners \cite{Jones2008, Chun2014}. Being able to confidently predict future trends of diseases, may lead to better anticipation of healthcare systems and allow better decision making, which should consequently provide higher quality service to those who are in need of it.

As many diseases are related, graphical modeling of disease data may be beneficial for predicting disease trends. As such, several types of graphs may be built and several prediction tasks may be imposed on these graphs. In the literature, several networks have been constructed to study the connection between human diseases. The nodes of the networks are disease codes while the links are derived based on common genes \cite{Goh2007}, shared metabolic pathways \cite{Lee2008}, connection to common miRNA molecule \cite{Lu2008}, observed comorbidity \cite{Hidalgo2009}. These networks have been used for the study of obesity \cite{Barabasi2011}, illness progression \cite{Hidalgo2009}, association of diseases with the cellular interactions \cite{Park2009}, properties of the genetic origin of diseases \cite{Goh2007}, etc. 

Since the HCUP are EHR data, we are more interested in modeling of phenotypic networks. These kind of networks are introduced in \cite{Davis2011, Hidalgo2009}. In \cite{Davis2011} a novel multi-relational link prediction method is proposed and it is shown that disease co-morbidity can enhance the current knowledge of genetic association.
In our study, we are primarily concentrated on disease-based graphs (we have opted modeling 253 diseases coded with CCS schema rather than modeling diseases with ICD-9 codes, because the CCS is the empirically built schema interpretable by wider audience rather than medical experts). Specifically, disease-based graphs can be built as comorbidity graphs based on phenotypic patient data (as described in Section~\ref{sec:commorbidity_graph}), but also other disease links based on common genes \cite{Davis2011}, ontologies, common patient profile, or common history.



As mentioned before, in previous work on diseases data, many graphs were proposed, and we now aim to utilize such constructed graphs to improve predictive power of unstructured models, which to our knowledge was not done before. For this task we are using a Gaussian Conditional Random Fields (GCRF) model. The GCRF model \cite{Radosavljevic2010} has been successfully applied in many real world problems, providing improvements over given state-of-the-art models while maintaining reasonable scalability. Some of the applications are in the climatology domain \cite{Djuric2014, Radosavljevic2010, Stojanovic2015}, modeling patient's response in acute inflammation treatment \cite{Radosavljevic2013}, traffic estimation \cite{Djuric2011}, image de--noising problems \cite{ristovski2013}, etc.

When models provide prediction for real-world problems, it can be very important to report uncertainty estimation of the prediction. This is especially true for domains where predictions are used for important decision-making, such as health. The objective of this paper is to improve the estimation quality of prediction uncertainty in the GCRF model for evolving graphs in order to more accurately represent the confidence of the prediction for healthcare applications. For example, instead of predicting that the number of admissions for a disease is going to be 15.026$\pm$10.000, making a different estimation of 15.000$\pm$150, can be more useful for decision making process. Therefore, we aim to address this important topic in this paper. We experimentally demonstrate improvement of predictive uncertainty by removing bias of the GCRF framework via representing its parameters as functions. We use two approaches, one that models parameters as a function of uncertainty estimation of unstructured predictors \cite{Radosavljevic2013}, (where we extend an initial approach as described in Section~\ref{sec:uGCRF}) and another one, that models parameters as neural networks \cite{Radosavljevic2014} (described in Section~\ref{sec:ufGCRF}).


In the following section, we are first going to describe how several unstructured models (Section~\ref{sec:unstructured}) and the GCRF method (Section~\ref{sec:gcrf_model}) allow modeling uncertainty, as well as how the uncertainty estimation can be improved with modeling parameters of the GCRF model as functions (Sections \ref{sec:uGCRF} and \ref{sec:ufGCRF}). Further, we are going to describe in more details the data we used in our experiments and the way of constructing the graphs from this data (Section~\ref{sec:data}). Experiments are described in Section \ref{sec:experiments} and results in terms of accuracy and uncertainty estimation quality are given in this same section. Finally, we are going to conclude this study and give some further goals in Section~\ref{sec:conclusion}.

\section{Models}

We aim to model complex trends of diseases in an autoregressive manner utilizing linear and non-linear models. These models will be called unstructured predictors as they have no information on graph structure, that might prove beneficial for predictive performance. Therefore, the structured GCRF framework will be used for introducing graph structure to these models.

\subsection{Unstructured Models}
\label{sec:unstructured}

For modeling disease trends we will be using several unstructured predictors, Linear regression models and Gaussian Processes regression models with several lags for learning each. These methods are evaluated and the ones with best performance are chosen as inputs for the GCRF framework, which will then introduce graph structure information to the prediction, providing a certain level of correction.
As uncertainty of the unstructured predictors is the additional information we are using in some of our experiments, we will provide formulas for retrieving uncertainty from unstructured predictors.

\subsubsection{Linear AR Model}

Linear regression form of auto--regressive (AR) representation is:
\begin{equation}
y_i = w^{T}x_{i} + \varepsilon , \varepsilon \sim \mathcal{N}(0, \sigma_{y}^{2})
\end{equation}
where $w$ is an unknown set of weights. The weight and noise variance are estimated by
\begin{equation}
\hat{w} = X^{T}(X^{T}X)^{-1}Xy .
\end{equation}
The variance of the Linear predictor is given by
\begin{equation}
\sigma_{y}^{2} = \frac{(y - \hat{w}^{T}X)^{T}(y - \hat{w}^{T}X)}{N - k - 1},
\end{equation}
where X is matrix representation of all data available for training, N is the number of training examples and k is the number of attributes. Prediction $y_{*}$ and uncertainty estimation $\sigma_{*}^{2}$ of the test data $x_{*}$ are found using
\begin{equation}
y_{*} = \hat{w}^{T}x_{*},
\end{equation}
\begin{equation}
\sigma^{2}_{*} = \sigma^{2}_{y}(1 + x_{*}(X^{T}X)^{-1}x_{*}^{T}).
\end{equation}

\subsubsection{Gaussian Processes Regression Model}

A Gaussian process (GP) is generalization of a multivariate Gaussian distribution over finite vector space to a function space of infinite dimension \cite{Rasmussen2006}. Assumption of a Gaussian prior is present over functions that map $x$ to $y$: 
\begin{small}
\begin{equation}
\label{eq:norm_assumption}
y_{i} = f(x_i) + \varepsilon, \varepsilon \sim \mathcal{N}(0, \sigma_{y}^{2}).
\end{equation}
\end{small}
Gaussian processes are defined with
\begin{small}
\begin{equation}
f(x) \sim GP(m(x),k(x,x')),
\end{equation}
\end{small}
where $m(x)$ is the mean function and $k(x, x')$ is the covariance function in the form of a kernel that is required to be positive definite.
In our implementation we are using a Gaussian kernel:
\begin{small}
\begin{equation}
k(x_i , x_j ) = \sigma_{y}^{2} exp \left( -\frac{1}{2} \sum_{d=1}^{D}\frac{(x_d^{i} - x_{d}^{j})^{2}}{w_{d}^{2}}. \right)
\end{equation}
\end{small}
Here, $ \sigma_{y}^{2}$ is the white noise variance parameter of the Gaussian prior. 

If we denote covariance of training part as $C = K+\sigma^{2}_{y}I_N$, $K_{ij} = k(x_i,x_j)$, the joint density of the observed outputs $\mathbf{y}$ and test output $y_{*}$ is presented as
\begin{small}
\begin{equation}
\binom{\mathbf{y}}{y_{*}} = \mathcal{N}\left(0, \begin{bmatrix}
C & \mathbf{k_{*}}\\ 
\mathbf{k_{*}^{T}} & c_{*} 
\end{bmatrix} \right),
\end{equation}
\end{small}
where $\mathbf{k_{*}}$ is the covariance vector for the new test point $\mathbf{x_{*}}$. The posterior predictive density is given by
\begin{small}
\begin{equation}
p(y_* | \mathbf{x_*}, \mathbf{X}, \mathbf{y}) = \mathcal{N}(y | 0, C),
\end{equation}
\end{small}
\vspace{-10pt}
\begin{small}
\begin{equation}
\mu_* = \mathbf{k_*^T}C^{-1}\mathbf{y}, \hspace{20pt} \sigma_{*}^{2} = c_* - \mathbf{k_{*}^{T}}C^{-1}\mathbf{k_*}.
\end{equation}
\end{small}
The $\sigma^2_*$ is the predictive variance or uncertainty at test point $x_{*}$ \cite{Rasmussen2006}.

\subsection{Structured Models}
Benefits of using structured methods have emerged in the previous decade, and there are many evidences of benefits of using structured methods over unstructured \cite{Radosavljevic2010}. As many datasets can be presented as graphs these methods have gained popularity and are considered state-of-the art methods in many domains.

\subsubsection{Continuous Conditional Random Fields}
Continuous Conditional Random Fields were proposed by \cite{Qin2008}, and model conditional distribution:
\begin{small}
\begin{equation}
P(y|X) = \frac{1}{Z(X, \mathbf{\alpha}, \mathbf{\beta})} exp(\phi(y,X,\mathbf{\alpha},\mathbf{\beta})).
\end{equation}
\end{small}
The term in the exponent $\phi(y,X,\mathbf{\alpha},\mathbf{\beta})$ and the normalization constant $Z(X, \mathbf{\alpha}, \mathbf{\beta})$ are defined as follows
\begin{small}
\begin{equation}
\phi(y,X,\mathbf{\alpha},\mathbf{\beta}) = \sum_{i=1}^{N} A(\mathbf{\alpha}, y_i, X) + \sum_{i \sim j} I(\mathbf{\beta}, y_i, y_j, x), 
\end{equation}
\end{small}
\vspace{-16pt}
\begin{equation}
Z(X, \mathbf{\alpha}, \mathbf{\beta}) = \int_{y} exp(\phi(y,X,\mathbf{\alpha},\mathbf{\beta})) d_y. 
\end{equation}
Function $A$ is called the association function, and it represents any function that handles mapping from $X \rightarrow y$ with respect to input variables $X$. Function $I$ is called interaction potential and it handles any relation the two data instances $y_i$ and $y_j$ have. In order to efficiently define the CRF form, association and interaction potentials are defined as linear combinations of feature functions $f$ and $g$ \cite{Lafferty2001},\begin{equation}
A(\mathbf{\alpha}, y_i, X) = \sum_{k=1}^{K} \mathbf{\alpha_{k}}f_{k}(y_i, X),
\end{equation}
\vspace{-15pt}
\begin{small}
\begin{equation}
I(\mathbf{\beta}, y_i, y_j, x) = \sum_{l=1}^{L} \mathbf{\beta_{l}}g_{l}(y_i, y_j, X).
\end{equation}
\end{small}
\vspace{-20pt}
\subsubsection{Gaussian Conditional Random Fields}
\label{sec:gcrf_model}

Feature function $f_{k}$ can be represented as the residual error of any unstructured predictor $R_{k}$, 
\begin{small}
\begin{equation}
f_{k}(y_i, X) = -(y_i - R_{k}(X))^{2}, k = 1,...K.
\end{equation}
\end{small}
And feature function $g_{l}$ as the residual error of given network similarity $S_{ij}^{(l)}$
\begin{small}
\begin{equation}
g_{l}(y_i, y_j, x) = -S_{ij}^{(l)}(y_i - y_j)^{2}, l = 1,...L.
\end{equation}
\end{small}
Then, the final GCRF takes following log-linear form:
\begin{small}
\begin{multline*}
P(y| X)= \frac{1}{Z}exp(-\sum_{i=1}^{K} \sum_{k=1}^{K}\ 
\alpha _{k} (y_i - R_k(X))^2 \\
  - \sum_{i \sim j} \sum_{l=1}^{
L}\beta_l {S_{ij}}^{(l)}(y_i - y_j)^2)
\end{multline*}
\end{small}
where $\alpha$ and $\beta$ are parameters of the feature functions, which model the association of each $y_i$ and $X$, and the interaction between different $y_i$ and $y_j$ in the graph, respectively. Here $R_k(x)$ functions are any unstructured predictors that map $X\rightarrow y_i$ independently, and might be used to also incorporate domain specific models. Similarity matrix $S$ is used to define the weighted undirected graph structure between labels.

This choice of feature functions enables us to represent this distribution as a multivariate Gaussian \cite{Radosavljevic2010} to ensure efficient and convex optimization:
\begin{small}
\begin{equation}
P(y|X) = \frac{1}{(2\pi)^{\frac{N}{2}}\mid \Sigma\mid^\frac{1}{2}}exp\left(-\frac{1}{2}(y - \mu)^{T}Q(y - \mu) \right )
\end{equation}
\end{small}
where $Q$ represents the inverse covariance matrix:
\begin{small}
\begin{equation}
Q = \left\{\begin{matrix}
2\sum_{k=1}^{K}\alpha_{k} + 2\sum_{k}\sum_{l=1}^{L}\beta_l e_{ij}^{(l)}S_{ij}^{(l)}(x) , i=j 
\\
2\- \sum_{l=1}^{L}\beta_l e_{ij}^{(l)}S_{ij}^{(l)}(x) , i\neq j
\end{matrix}\right.
\end{equation}
\end{small}
The posterior mean is given by 
\begin{small}
\begin{equation}
\mu = Q^{-1} \bf b,
\end{equation}
\end{small}
where $b$ is defined as
\begin{small}
\begin{equation}
b_i = 2\left(\sum_{k=1}^{K}\alpha_{k} R_{k}(x) \right ).
\end{equation}
\end{small}
This specific way of modeling will allow efficient inference and learning. It should be noted that the GCRF model does not depend on input variables $X$ directly, but via its inputs $R_{k}(X)$ and $S_{ij}^{(l)}$ which are learned prior to GCRF learning. The parameters $\alpha$ and $\beta$ serve as confidence indicators, however are not sensitive to any change of distribution of input variables, which is common in real life datasets, thus making them biased towards the unstructured predictors whose performances may vary on the dataset. In this paper we aim to address this bias problem of the GCRF model, the extensions are going to be described in Sections~\ref{sec:uGCRF} and \ref{sec:ufGCRF}.

\paragraph{Learning and inference}

The learning task is to optimize parameters $\alpha$ and $\beta$ by maximizing the conditional log-likelihood,
\begin{small}
\begin{equation}
(\hat{\alpha }, \hat{\beta}) = \underbrace{argmax}_{\alpha, \beta}logP(y|X;\alpha,\beta)
\end{equation}
\end{small}
which is a convex objective, and can be optimized using standard quasi-newton optimization techniques. Note that there is one constraint that is needed to assure the distribution is Gaussian, which is to make the $Q$ matrix positive-semidefinite. To ensure this, we are using the exponential transformation of parameters $\alpha_{k} = e^{u_{k}}$ and $\beta_{l} = e^{v_{l}}$, as suggested in \cite{Qin2008}, to make the optimization unconstrained.

\subsection{Parameters of GCRF as functions}
\label{sec:uncertainty_functions}

The GCRF intrinsically possesses uncertainty estimation. This uncertainty estimation is highly biased towards the data the model was trained on as GCRF does not depend on input variables directly.
Once parameters of the GCRF model are introduced as functions, these functions impact both parameters values and scale (note that $\sum_k \alpha_k + \sum_l \beta_l = I$, where $I$ is a unit vector). Parameters in the GCRF represent degree of belief toward each unstructured predictor and similarity measure. If they are modeled to be dependent on input variables, the bias problem will be solved and thus uncertainty estimation should be improved. The uncertainty estimation improvement is achieved both by better fitting parameters and by altering the unit scale to better fit the data as well.
We experimentally evaluate our assumptions on modeling parameters of the GCRF model as functions in terms of uncertainty estimation and prediction accuracy (Section~\ref{sec:experiments}).
In this chapter we summarize potential ways of handling the bias in the GCRF models caused from parameters $\alpha_k$.

\subsubsection{Parameters $\alpha_{k}$ as functions of unstructured predictor's uncertainty}
\label{sec:uGCRF}

First, we address the problem of model bias by modeling the overall model uncertainty as a function of the uncertainty of each unstructured prediction. The principal assumption is that the chosen unstructured predictors can output uncertainty for their predictions. 
Initial GCRF uncertainty estimation improvements were done on modeling time series of patients' response to acute inflammation treatment \cite{Radosavljevic2013} showing that the uncertainty estimation of GCRF modeled in this way provides a higher quality of uncertainty than the quality of this measure for utilized unstructured predictors. The new parameters of the GCRF model are modeled such that:
\begin{small}
\begin{equation}
\alpha_{k,p}=\frac{e^{u_{k,p}}}{\sigma_{k, 1}^{2}} , \beta=e^{v}
\end{equation}
\end{small}
where $\sigma_{k, 1}^2$ represents the uncertainty estimation of unstructured predictor k for the first time step ($p=1$), while $\alpha$ and $\beta$ are coefficients for the feature functions. 

We have extended this work by relaxing a few assumptions made in \cite{Radosavljevic2013}, as well as by applying uncertainty estimation on evolving graphs data (rather than applying it on linear--chain data). The first assumption of the previous work was that the log likelihood of GCRF would be optimized with respect to the uncertainty of each unstructured predictor, but only considering the first time step of the respective model. This follows the homoscedasticity assumption that the variance of the model will not change significantly through time. This strong assumption of homoscedasticity is dropped in our study and the parameters of the GCRF model have been optimized with respect to the uncertainty estimation of each unstructured predictor at each time step. This allows the model to optimize different parameters for different prediction horizons. Thus, if predictions of the unstructured predictor increase uncertainty further in the future, GCRF will also adjust accordingly. 

We have further improved uncertainty estimation by penalizing the predictors whose uncertainty estimation was not good enough during validation. This could be done by any quality measure of uncertainty estimation on the training data. We use the percentage of nodes that fall into the $95\%$ confidence interval ($ci_{k,p}$) as a quality index to augment our approach:
\begin{small}
\begin{equation}
\alpha_{k,p}=\frac{e^{u_{k,p}}}{\sigma_{k, p}^{2}}ci_{k,p}.
\end{equation}
\end{small}
\vspace{-15pt}
\subsubsection{Parameters $\alpha_{k}$ as functions of input variables}
\label{sec:ufGCRF}

The previous approach can be further generalized by observing parameters $\alpha$ as functions of input parameters, which was proposed in \cite{Radosavljevic2014}. However, in \cite{Radosavljevic2014} there were no experimental results on uncertainty improvement and the paper does not mention this aspect of the extension. We can observe the parameter $\alpha$ as a parameterized function of input variables $\alpha(\theta_{k},x)$, where $\theta_{k}$ are the parameters, and $x$ are input variables of function $\alpha$. Due to the positive semi--definiteness of the precision matrix constraint, function $\alpha(\theta_{k},x)$  becomes
\begin{small}
\begin{equation}
\alpha_{k}= e^{u_{k}(\theta_{k},x)} , 
\end{equation}
\end{small}
where $u_{k}(\theta_{k},x)$ is in \cite{Radosavljevic2014} a feed-forward neural network. This method is easily optimized using gradient descent methods. These functions are titled as uncertainty functions, since they provide significant improvement to the covariance matrix in terms of bias correction.

\section{Medical Data}
\label{sec:data}

The data source we used in this study, as we mentioned previously, is the State Inpatient Databases  (SID) which is an archive that stores the universe of inpatient discharge abstracts from data organizations. It is provided by the Agency for Healthcare Research and Quality and is included in the Healthcare Cost and Utilization Project (HCUP) \cite{hcup}. Particularly, we have access to the SID California database, which contains 35.844.800 inpatient discharge records over 9 years (from January 2003 to December 2011) for 19.319.350 distinct patients in 474 hospitals (436 AHAID identified; about 400 per year). For each patient there are up to 25 diagnosis codes in both CSS and ICD9 coding schemas, together with up to 25 procedures applied during this particular admission of the patient. There are also some demographic information about each patient, like age, birth year, sex, race, etc., as well as information about hospital stays, length of stay, total charges, type of payment, a payer, discharge month, survival information. 

Availability of this information made possible building of several graphs, and exploring usefulness of attributes and links of the graph for a chosen target attribute. In this study, we focused on disease based graphs and prediction of number of admissions for each disease. Graphs are constructed for these 9 years in monthly resolution. In our experiments, we used 231 out of 259 diseases that we were able to follow throughout all 9 years. For each node,  we include temporal information, meaning that there are one, two or three previous values of the target variable (we refer to these attributes as lag1, lag2, and lag3) as attributes of a node in the current time step (more details about utilization of those attributes are given in Section~\ref{sec:experiments}). 
When it comes to the structure, we explored several cases. 
\subsection{Comorbidity graph}
\label{sec:commorbidity_graph}
\begin{figure}[h!]
\vspace{-5pt}
\includegraphics[scale=0.17]{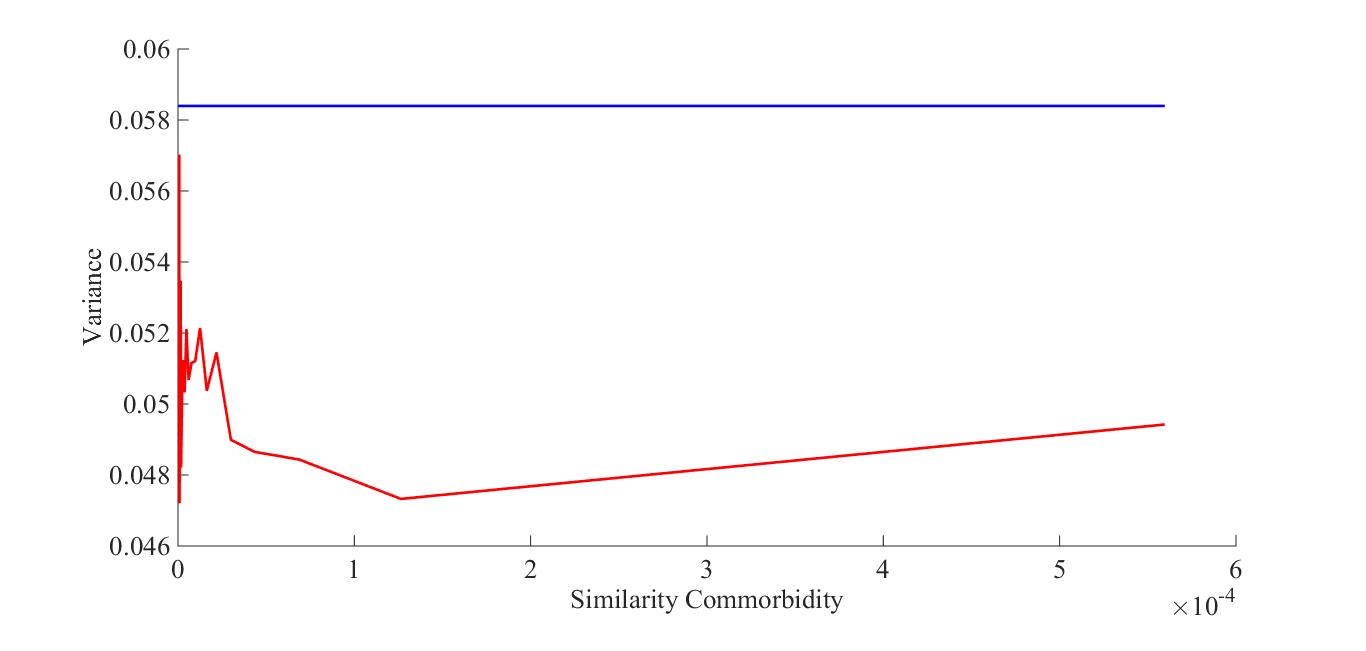}
\caption{Comorbidity graph variogram. Blue line represents disease comorbidity similarity and red line represents overall variance.}
\vspace{-10pt}
\label{fig:sim_commorbidity}
\end{figure}

Disease-based graphs can be built as comorbidity graphs based on phenotypic patient data \cite{Davis2011}. However, this type of graph did not provide satisfactorily results, since variogram \cite{Uversky2014} for the selected target attributes was not appropriate (Figure~\ref{fig:sim_commorbidity}). Variograms were inspected for two behaviors: displaying a decreasing trend in variance at higher values of similarity, and falling below the line of the overall variance of the data. By \cite{Uversky2013} this similarity measure would be characterized as a ''bad'' similarity measure. Therefore, we researched several other disease links, like those based on common patient profile or common history. 
\vspace{-10pt}
\subsection{Jenson-Shannon divergence graph}
\label{sec:js_graph}
\begin{figure}[h!]
\includegraphics[scale=0.17]{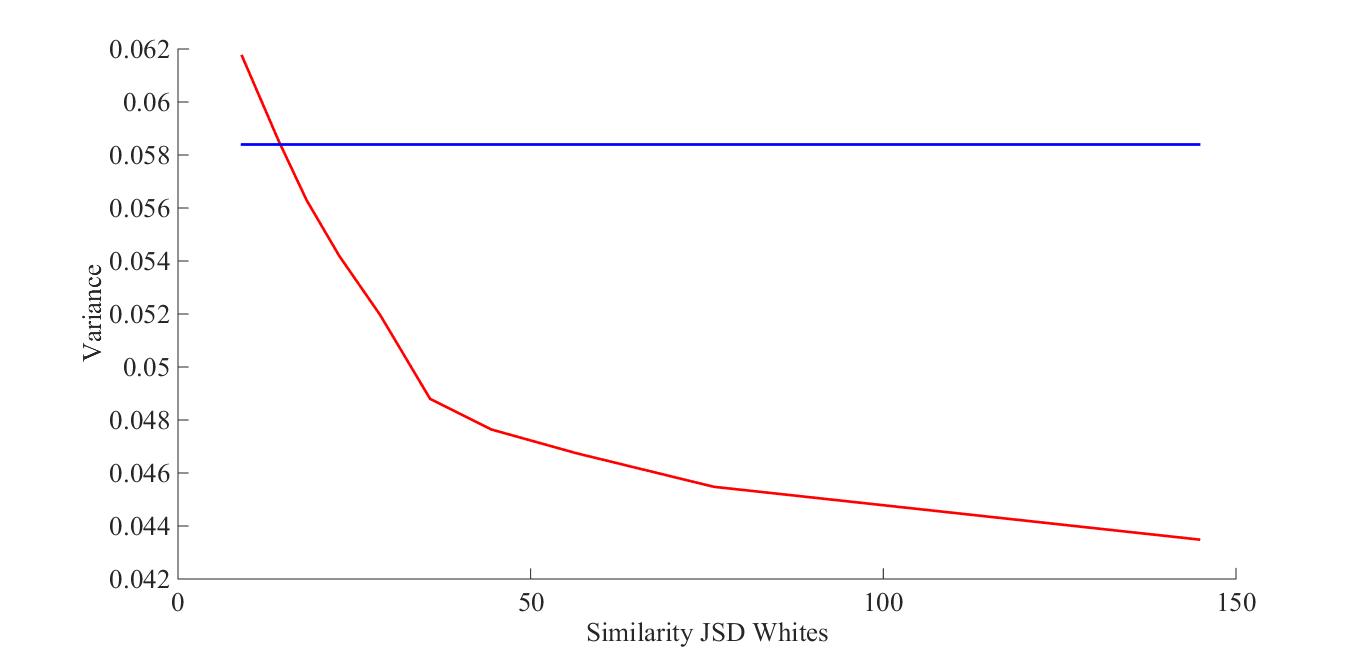}
\caption{JS divergence graph variogram. Blue line represents disease similarity according to JS divergence and red line represents overall variance.}
\vspace{-10pt}
\label{fig:sim_js}
\end{figure}

For the first case we measured Janson-Shanon divergence between distribution of two diseases based on several attributes.
\begin{small}
\begin{equation}
JSD(P || Q)=\frac{1}{2}(KLD(P || M)+KLD(Q || M)) ,
\end{equation}
\end{small}
where $KLD$ is the Kullback-Leibler divergence, $P$ and $Q$ are the distributions of the selected disease attribute for two observed diseases and $M = \frac{1}{2}(P+Q)$. 
The similarity obtained from this divergence is 
\begin{small}
\begin{equation}
S(y_p, y_q) = \frac{1}{JSD(x_p || x_q)}.
\end{equation}
\end{small}
Utilizing the variogram technique showed that using the distribution of white people admitted for each disease showed the best performance among all other attributes of each disease (Figure~\ref{fig:sim_js}). 
\subsection{Common history graph}
\label{sec:hist_graph}
\begin{figure}[h!]
\includegraphics[scale=0.17]{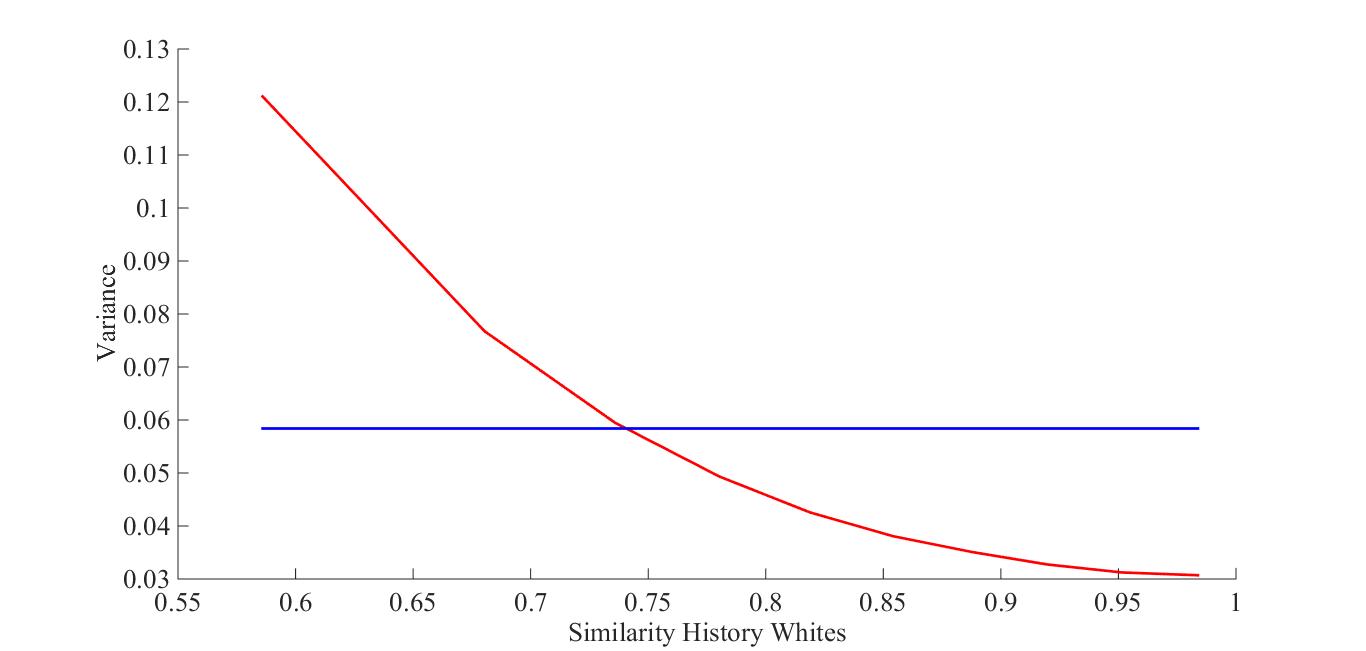}
\caption{Similarity history variogram. Blue line represents disease history similarity and red line is overall variance.}
\vspace{-5pt}
\label{fig:sim_history}
\end{figure}

The best performance so far was obtained by the structure built on common history links as can be seen in Figure~\ref{fig:sim_history}, calculated using formula:
\begin{equation}
S(y_i, y_j) = exp ( - mean(abs(x_i^{h} - x_j^{h}))).
\end{equation}
Here, $x_i^{h}$ represents vector of length $h$ utilized for given attribute of the node $x$. For example, if we were \begin{small}
observing previous 2 time steps the similarity would be 
\begin{equation}
S(y_i, y_j) = exp ( - \frac{abs((x_i^{t1} - x_j^{t1}) + (x_i^{t2} - x_j^{t2}))}{2}).
\end{equation}
\end{small}
The variogram was obtained using historical information of admitted whites for each observed disease in previous three timesteps.  
As this variogram is the best, we have decided to use this measure in our experiments.

\section{Experiments}
\label{sec:experiments}
We have characterized mentioned methodology approaches to the HCUP diseases evolving graph. Our goal is to predict the number of people admitted for a disease as primary one in twelve months of the year 2011 in the state of California, which is the last year in our database.
We have normalized the admission count to a 0-1 scale, making admission rate as the target variable. Predicted values can easily be converted back to counts of people admitted for that disease. Between nodes in each timestep we have defined the similarity values based on the exploration in Section~\ref{sec:data}.

For the unstructured predictors we have used a linear and non-linear model. Both models were trained in an autoregressive fashion for each disease separately, observing several previous timesteps to infer one timestep ahead. For each of the two models we have used target variables of up to three timesteps in history (lags) as features of the models. Unstructured predictors were optimized with a sliding window of 12 months.
Among the unstructured predictors, the best linear and non-linear predictors were chosen as inputs into the GCRF model.
Separate GCRF models were optimized with linear and non-linear predictors so the effect of each on the results could be characterized separately. 

We are training and comparing three different GCRF models. The first is an original GCRF model (described in Section~\ref{sec:gcrf_model}), with a slight difference that an $\alpha$ parameter was optimized for each node in a graph separately for fair comparison with the other GCRF models. We will denote this model simply as GCRF. 
The second GCRF model is the one described in Section~\ref{sec:uGCRF}, which includes the uncertainty estimation of the unstructured predictor to rescale parameter $\alpha$ with uncertainty of the unstructured predictor. In our results this method is called uGCRF. 
Finally, the third GCRF model is the one described in Section~\ref{sec:ufGCRF}, where parameters $\alpha$ were modeled as feed-forward neural networks. The last GCRF model, with uncertainty functions, will be denoted as ufGCRF in our results.

We will characterize all methods for their predictive power using the root mean squared error (RMSE). As we have mentioned before, an important goal of mentioned approaches is the improved uncertainty estimation power, which will be evaluated using negative log predictive density as we describe below.

\paragraph{Uncertainty estimation quality measure}

The quality of uncertainty estimates is evaluated by measuring the average Negative Log Predictive Density (NLPD), a measure that incorporates both the accuracy and uncertainty estimate of the model. NLPD metric penalizes both over and under confident predictions. This measure is also used in data analysis competitions \cite{Quinonero-Candela2006a} where uncertainty estimation was of major importance. Smaller values of NLPD correspond to better quality of the estimates. For given prediction $y_{i*}$, NLPD reaches a minimum for $\sigma_{i*}^{2}=(y_{i}-y_{i*})^{2}$.
\begin{equation}
NLPD=\frac{1}{2} \sum_{i=1}^{N} \frac{(y_{i}-y_{i*})^{2}}{2\sigma_{i*}^{2}}+log\sigma_{i*}^{2}
\end{equation}
where $\sigma_{i*}^{2}$ is predicted uncertainty.

\paragraph{Experimental results}
In our experiments, unstructured predictors provide higher predictive accuracy when using only the previous timestep as input (lag 1). As such, we have used the Linear regression and Gaussian Processes regression with lag 1 as unstructured models for the structured GCRF model. Results are shown in Table~\ref{tab:rmse_results}, the best results are bolded, the runner--up results are underlined, and third best results are in italic. 
\begin{figure}[h!]
\includegraphics[scale=0.65]{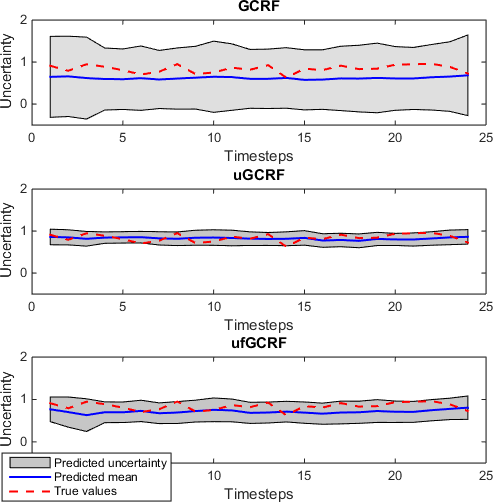}
\caption{Confidence interval plots for the three GCRF methods. Blue lines represent true values of admission rate for Hepatitis disease (24 months for years 2010 and 2011), while red lines represent prediction of the respective GCRF.}
\vspace{-22pt}
\label{fig:unc_gcrf}
\end{figure}
From the Table~\ref{tab:rmse_results} it can be observed that the GCRF model using GP as unstructured predictor gives the most accurate result and that two other GCRF models introduce slightly more error to the original GCRF model. The main objective of uncertainty functions is to scale belief towards unstructured predictors with respect to change of their respective input variables. Rescaling variance values (from which uncertainty is expressed) is another, more robust effect of uncertainty functions that significantly benefits the models uncertainty estimation, as described in Section~\ref{sec:uncertainty_functions}.
Following the original GCRF method results, the ufGCRF model introduces less error than the uGCRF method deeming it superior among the two extensions in this experiment. It should be noted that all GCRF methods displayed outperform all of the unstructured predictors. This evidently comes from the structure information included in the prediction. As a reminder, we are using common history graph described in Section~\ref{sec:hist_graph}.

The other aspect of characterizing the methods includes evaluation of uncertainty quality, which is an important concept in this study. Uncertainty estimation is evaluated using the NLPD metric and the results are shown in the Table~\ref{tab:nlpd_results}. The original GCRF model provides lowest quality of the uncertainty estimation among all of the other methods. This can be attributed to the underconfident predictions of the GCRF model, which can be observed also in Figure~\ref{fig:unc_gcrf}. By $underconfidence$ we refer to the estimated confidence interval being too high (large gray area in Figure~\ref{fig:unc_gcrf} in the case of the original GCRF). For instance in Figure~\ref{fig:unc_gcrf} we observe that GCRF is giving estimate for admission of Hepatitis of about 0.61$\pm$0.78, making predictions of the GCRF model less useful. The two extensions narrow down the unnecessary large confidence interval in the original GCRF case, where their predictions are 0.81$\pm$0.16 for uGCRF and 0.71$\pm$0.25 for ufGCRF, which can be more helpful for decision making process.
Since NLPD measure takes into account both accuracy of the model and uncertainty estimation quality, we see that this ability to narrow confidence interval of the two extensions, is also reflected in the results shown in Table~\ref{tab:nlpd_results}. The fact that ufGCRF has better predictive power gives it better results in NLPD than uGCRF, which shows good uncertainty estimation, but introduce little bit more error than ufGCRF. Actually, the best performance in terms of NLPD measure gives ufGCRF using LR as unstructured predictor, but right after it goes again ufGCRF with GP, so we can conclude that this GCRF extension gives the best uncertainty estimation in our experiments among all models.
\begin{figure}[h!]
\vspace{-3pt}
\begin{center}
\subfigure[GCRF]{\includegraphics[width=83mm, height = 20mm]{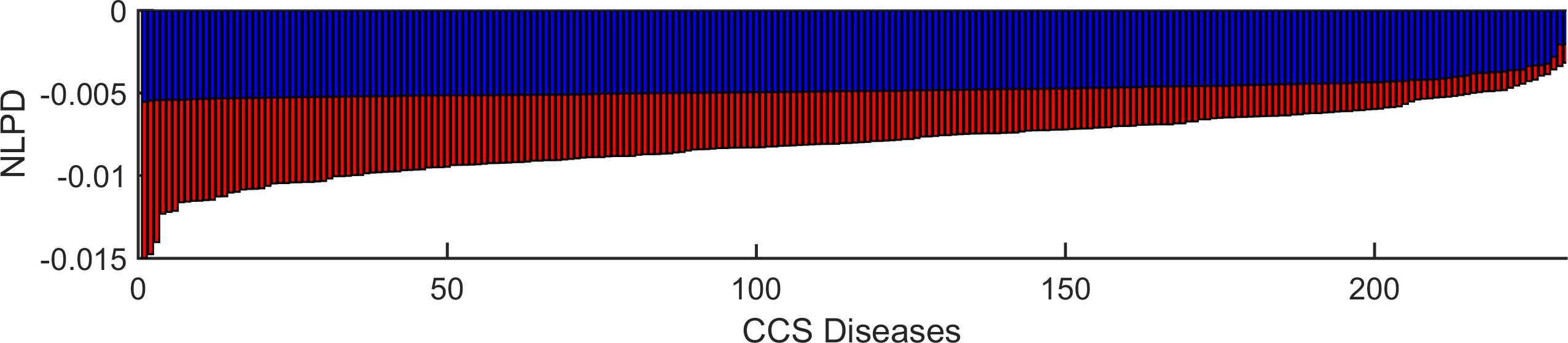}}
\subfigure[uGCRF]{\includegraphics[width=83mm, height = 20mm]{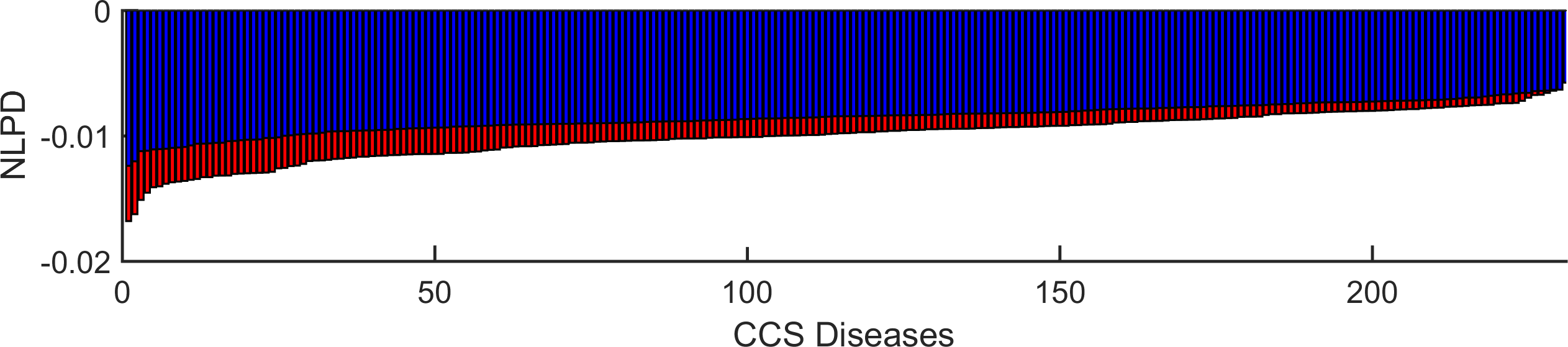}}
\subfigure[ufGCRF]{\includegraphics[width=83mm, height = 20mm]{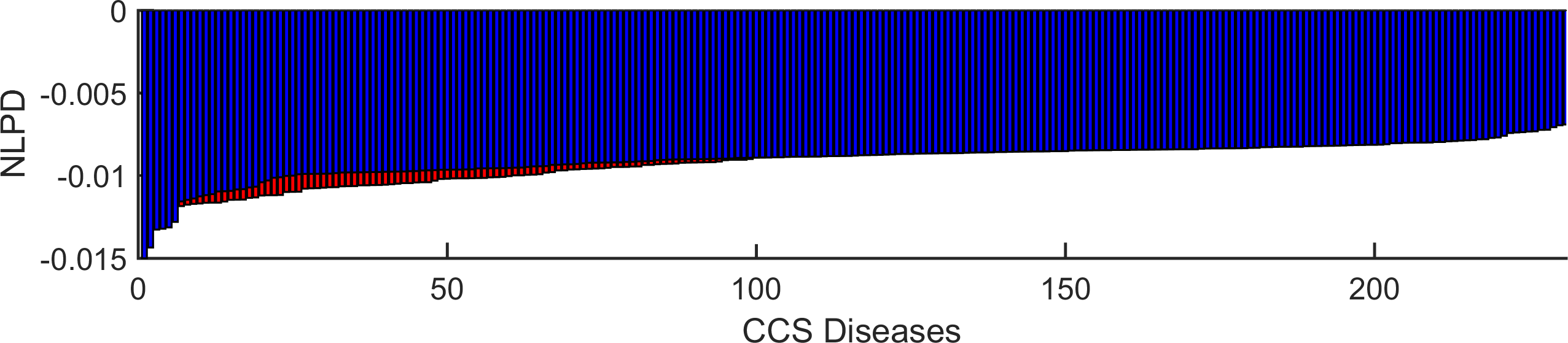}}
\end{center}
\vspace{-15pt}
\caption{\small Min and obtained NLPD (y-axis; smaller is better) for all 231 diseases (x--axis) for three GCRF models with Linear Regression as unstructured predictor. Red bars represent the average \textbf{minimal NLPD} value for given prediction of the node over 12 months and blue bars are the average \textbf{obtained NLPD} value for that node over 12 months.}%
\label{fig:nlpd_gcrf_lr}
\vspace{-10pt}
\end{figure}
The comparison of uncertainty estimation quality of the three GCRF methods is also given in Figure~\ref{fig:nlpd_gcrf_lr} (this figure shows result using LR as unstructured predictor, however results for using GP are similar and are omitted for lack of space). Each bar represents mean NLPD measure for each disease averaged over the 12--months testing period. Red bars represent minimal NLPD value for given predictions (the best possible uncertainty estimation), while blue bars represent actual uncertainty quality obtained for NLPD metric in experiments. Values are sorted in descending order to show rate of the blue and red surface, where the model that has more dominant blue surface is better. We can see that GCRF fails to reach the minimal NLPD for its predictions, while uGCRF does a much better job. However, the best performance is provided by ufGCRF, where we see that obtained NLPD is very close to the minimal NLPD for given predictions in most cases. Both uGCRF and ufGCRF appear to estimate uncertainty much closer to the true error variance, with ufGCRF having superior performance in both uncertainty estimation and predictive power among two extensions. Also, note that both uGCRF and ufGCRF outperform the GCRF model for \textit{each} disease in terms of uncertainty estimation.

\begin{table*}[t]
  \centering
  \caption{RMSE results of admission rate prediction. First column are averaged results over 12 months of prediction period and the following columns are prediction results for each month in the year 2011. The best results are bolded, second best are underlined and third best are in italic.}
  \resizebox{\textwidth}{!}{%
    \begin{tabular}{|c|c|clllllllllll|}
    \toprule
    \textbf{RMSE}  & \textbf{Average} & January & February & March & April  & May   & June  & July  & August & September & October & November & December \\
    \midrule
    LR lag1 & 0.3168 & 0.3027 & 0.3267 & 0.3504 & 0.3083 & 0.3004 & 0.3153 & 0.3119 & 0.3196 & 0.3154 & 0.3132 & 0.3142 & 0.3230 \\
    LR lag2 & 0.3296 & 0.3236 & 0.3262 & 0.3644 & 0.3191 & 0.3051 & 0.3228 & 0.3230 & 0.3331 & 0.3341 & 0.3320 & 0.3326 & 0.3393 \\
    LR lag3 & 0.3536 & 0.3483 & 0.3476 & 0.3967 & 0.3509 & 0.3312 & 0.3553 & 0.3465 & 0.3487 & 0.3588 & 0.3490 & 0.3535 & 0.3566 \\
    \hline
    GP lag1 & 0.3098 & 0.3013 & 0.3306 & 0.3309 & 0.2933 & 0.2935 & 0.3097 & 0.3063 & 0.3176 & 0.3124 & 0.3035 & 0.2988 & 0.3200 \\
    GP lag2 & 0.3129 & 0.3073 & 0.3281 & 0.3338 & 0.2940 & 0.2952 & 0.3132 & 0.3127 & 0.3221 & 0.3141 & 0.3071 & 0.3029 & 0.3236 \\
    GP lag3 & 0.3158 & 0.3092 & 0.3299 & 0.3354 & 0.2961 & 0.2949 & 0.3182 & 0.3172 & 0.3239 & 0.3160 & 0.3092 & 0.3061 & 0.3330 \\
    \hline
    GCRF + LR & \underline{0.2486} & \underline{0.2372} & \underline{0.2769} & \underline{0.2795} & \underline{0.2328} & \underline{0.2343} & \underline{0.2486} & \underline{0.2464} & \underline{0.2509} & \underline{0.2449} & \underline{0.2434} & \underline{0.2435} & \underline{0.2445} \\
    uGCRF + LR & 0.3072 & 0.2938 & 0.3193 & 0.3395 & 0.2956 & 0.2900 & 0.3073 & 0.3025 & 0.3106 & 0.3061 & 0.3041 & 0.3049 & 0.3123 \\
    ufGCRF + LR & 0.2713 & \textit{0.2540} & \textit{0.2785} & 0.3115 & 0.2523 & \textit{0.2526} & \textit{0.2709} & \textit{0.2662} & 0.2764 & 0.2745 & 0.2727 & 0.2746 & \textit{0.2716} \\
    \hline
    GCRF + GP & \textbf{0.2447} & \textbf{0.2365} & \textbf{0.2686} & \textbf{0.2717} & \textbf{0.2298} & \textbf{0.2325} & \textbf{0.2472} & \textbf{0.2460} & \textbf{0.2532} & \textbf{0.2415} & \textbf{0.2362} & \textbf{0.2320} & \textbf{0.2410} \\
    uGCRF + GP & 0.3013 & 0.2923 & 0.3211 & 0.3229 & 0.2840 & 0.2850 & 0.3027 & 0.2983 & 0.3100 & 0.3038 & 0.2953 & 0.2903 & 0.3102 \\
    ufGCRF + GP & \textit{0.2685} & 0.2569 & 0.2839 & \textit{0.2943} & \textit{0.2516} & 0.2532 & 0.2718 & 0.2669 & \textit{0.2762} & \textit{0.2697} & \textit{0.2635} & \textit{0.2610} & 0.2734 \\
    \bottomrule
    \end{tabular}}%
  \label{tab:rmse_results}%
\end{table*}%

\begin{table*}[t]
  \centering
  \caption{NLPD results of admission rate prediction. First column are averaged results over 12 months of prediction period and the following columns are prediction results for each month in the year 2011. The best results are bolded, second best are underlined and third best are in italic.}
  \resizebox{\textwidth}{!}{%
    \begin{tabular}{|c|c|clllllllllll|}
    \toprule
   \textbf{NLPD}  & \textbf{Average} & January & February & March & April  & May   & June  & July  & August & September & October & November & December \\
    \midrule
    LR lag0 & -0.0082 & -0.0085 & -0.0082 & -0.0082 & -0.0083 & -0.0084 & -0.0083 & -0.0083 & -0.0083 & -0.0083 & -0.0083 & -0.0083 & -0.0068 \\
    LR lag1 & -0.0080 & -0.0086 & -0.0082 & -0.0078 & -0.0075 & -0.0080 & -0.0082 & -0.0082 & -0.0080 & -0.0081 & -0.0080 & -0.0081 & -0.0068 \\
    LR lag2 & -0.0075 & -0.0085 & -0.0077 & -0.0071 & -0.0069 & -0.0071 & -0.0076 & -0.0077 & -0.0076 & -0.0076 & -0.0077 & -0.0077 & -0.0064 \\
    \hline
    GP lag0 & -0.0079 & -0.0081 & -0.0079 & -0.0080 & -0.0080 & -0.0080 & -0.0080 & -0.0080 & -0.0079 & -0.0080 & -0.0080 & -0.0080 & -0.0066 \\
    GP lag1 & -0.0079 & -0.0084 & -0.0080 & -0.0079 & -0.0079 & -0.0080 & -0.0080 & -0.0080 & -0.0079 & -0.0079 & -0.0079 & -0.0079 & -0.0067 \\
    GP lag2 & -0.0079 & -0.0086 & -0.0080 & -0.0078 & -0.0079 & -0.0080 & -0.0080 & -0.0080 & -0.0079 & -0.0079 & -0.0079 & -0.0079 & -0.0066 \\
    \hline
    GCRF + LR & -0.0044 & -0.0047 & -0.0044 & -0.0046 & -0.0046 & -0.0046 & -0.0046 & -0.0045 & -0.0044 & -0.0043 & -0.0043 & -0.0043 & -0.0038 \\
    uGCRF + LR & \textit{-0.0085} & \textit{-0.0088} & \textit{-0.0084} & \textit{-0.0086} & \textit{-0.0087} & \textit{-0.0088} & \textit{-0.0087} & \textit{-0.0087} & \textit{-0.0087} & \textit{-0.0087} & \textit{-0.0086} & \textit{-0.0086} & \textit{-0.0070} \\
    ufGCRF + LR & \textbf{-0.0090} & \textbf{-0.0094} & \textbf{-0.0086} & \textbf{-0.0090} & \textbf{-0.0093} & \textbf{-0.0093} & \textbf{-0.0095} & \textbf{-0.0092} & \textbf{-0.0092} & \textbf{-0.0095} & \textbf{-0.0093} & \textbf{-0.0092} & \textbf{-0.0066} \\
    \hline
    GCRF + GP & -0.0046 & -0.0048 & -0.0046 & -0.0047 & -0.0047 & -0.0048 & -0.0047 & -0.0046 & -0.0045 & -0.0045 & -0.0045 & -0.0045 & -0.0040 \\
    uGCRF + GP & -0.0083 & -0.0087 & -0.0083 & -0.0084 & -0.0086 & -0.0085 & -0.0085 & -0.0084 & -0.0084 & -0.0084 & -0.0084 & -0.0084 & -0.0070 \\
    ufGCRF + GP & \underline{-0.0089} & \underline{-0.0092} & \underline{-0.0083} & \underline{-0.0089} & \underline{-0.0091} & \underline{-0.0091} & \underline{-0.0093} & \underline{-0.0091} & \underline{-0.0090} & \underline{-0.0093} & \underline{-0.0092} & \underline{-0.0092} & \underline{-0.0067} \\
    \bottomrule
    \end{tabular}}%
  \label{tab:nlpd_results}%
\end{table*}%
\section{Conclusions and Future Work}
\label{sec:conclusion}
In this study, the GCRF model is successfully applied to a challenging problem of admission rate prediction, based on a temporal graph built from HCUP (SID) data. For sensitive health care and medical applications, one should consider aspects of uncertainty, and use this information in a decision making process. Thus, it was important to address this aspect of the methods and compare their quality of uncertainty estimation. 
In the experiments we characterize several unstructured (Linear Regression and Gaussian Processes with lag 1, lag 2 and lag 3) and structured predictors (original GCRF, uGCRF and ufGCRF) for their predictive error and quality of uncertainty estimation.  
All three structured models outperformed unstructured ones in terms of predictive error, showing that structure brings useful information to this prediction task. However in terms of quality of uncertainty estimation, those unstructured predictors did a better job than original GCRF, but underperformed compared to the two extensions of the GCRF model. 
Even though the original GCRF model showed the best performance in predictions, it had the lowest quality of uncertainty estimation. Introducing small predictive error, uGCRF and ufGCRF models gained large improvements in uncertainty estimation, especially the ufGCRF model that had the better performance in prediction of these two  GCRF model extensions.
As the next task in future work, we plan to introduce uncertainty propagation in the GCRF model as the predictive horizon increases.

\section{Acknowledgments}
This research was supported by DARPA Grant FA9550-12-1-0406 negotiated by AFOSR, National Science Foundation through major research instrumentation grant number CNS-09-58854. Healthcare Cost and Utilization Project (HCUP), Agency for Healthcare Research and Quality, provided data used in this study.

\bibliographystyle{siam}
\bibliography{SDM15HealthcareWorkshop}

\begin{thebibliography}{10}

\bibitem{hcup}
{\sc Agency for Healthcare Research and Quality}, {\em HCUP State Inpatient
  Databases (SID). Healthcare Cost and Utilization Project (HCUP)}, Rockville,
  MD, 2005-2009.

\bibitem{Barabasi2011}
{\sc A.L. Barabasi, N.~Gulbahce, and J.~Loscalzo}, {\em Network medicine: a
  network-based approach to human disease}, Nature Reviews Gen.,  (2011).

\bibitem{Davis2011}
{\sc D.A. Davis and N.~Chawla}, {\em Exploring and exploiting disease
  interactions from multi-relational gene and phenotype networks}, PLoS ONE,
  (2011).

\bibitem{Dey2014}
{\sc S.~Dey, J.S. Gyorgy, Westra B.L., M.~Steinbach, and V.~Kumar}, {\em Mining
  interpretable and predictive diagnosis codes from multi-source electronic
  health records}, in SDM, 2014.

\bibitem{Djuric2011}
{\sc N.~Djuric, V.~Radosavljevic, V.~Coric, and S.~Vucetic}, {\em Travel speed
  forecasting by means of continuous conditional random fields}, Journal of the
  Transportation Research,  (2011).

\bibitem{Djuric2014}
{\sc N.~Djuric, V.~Radosavljevic, Z.~Obradovic, and S.~Vucetic}, {\em Gaussian
  conditional random fields for aggregation of operational aerosol retrievals},
  Geosc. and Rem. Sens. Let.,  (2015).

\bibitem{Lee2008}
{\sc Lee D.S., Park J., Kay K.A., Christakis N.A., Oltvai Z.N., and Barabasi
  A.L.}, {\em The implications of human metabolic network topology for disease
  comorbidity}, Proc. of the Nat. Acad. of Sc. of the U.S.A., PNAS,  (2008).

\bibitem{Hidalgo2009}
{\sc C.A. Hidalgo, N.~Blumm, A.L. Barabasi, and N.~Christakis}, {\em A dynamic
  network approach for the study of human phenotypes}, PLoS Comput. Biol.,
  (2009).

\bibitem{Jones2008}
{\sc K.~Jones, N.~Patel, M.~Levy, A.~Storeygard, D.~Balk, J.~Gittleman, and
  P.~Daszak}, {\em Global trends in emerging infectious diseases}, Nature 451,
  (2008).

\bibitem{Goh2007}
{\sc Goh K.I., Cusick M.E., Valle D., Childs B., Vidal M., and Barabasi A.L.},
  {\em The human disease network}, Proc. of the Nat. Acad. of Sc. of the USA,
  PNAS,  (2007).

\bibitem{Chun2014}
{\sc Chun-Chi L., Yu-Ting T., Wenyuan L., Chia-Yu W., Ilya M., Andrey R.,
  Fengzhu S., Michael S.W., Jeremy C., Preet C., Joseph L., Edward C., and
  Xianghong Z.}, {\em Diseaseconnect: a comprehensive web server for
  mechanism-based disease-disease connections}, Nucl. Acids Res.,  (2014).

\bibitem{Lafferty2001}
{\sc J.~Lafferty, A.~McCallum, and F.~Pereira}, {\em Conditional random fields:
  Probabilistic models for segmenting and labeling sequence data}, in ICML,
  2001.

\bibitem{Lu2008}
{\sc M.~Lu, Q.~Zhang, M.~Deng, J.~Miao, Y.~Guo, W.~Gao, and Q.~Cui}, {\em An
  analysis of human microrna and disease associations}, PLoS ONE,  (2008).

\bibitem{Park2009}
{\sc J.~Park, D.S. Lee, N.A. Christakis, and A.L. Barabasi}, {\em The impact of
  cellular networks on disease comorbidity}, Mol. Syst. Biol.,  (2009).

\bibitem{Qin2008}
{\sc T.~Qin, T.Y. Liu, X.D. Zhang, D.S. Wang, and H.~Li}, {\em Global ranking
  using continuous conditional random fields}, in NIPS, 2008.

\bibitem{Quinonero-Candela2006a}
{\sc J.~Quinonero-Candela, C.~E. Rasmussen, F.~Sinz, and B.~Schoelkopf}, {\em
  Evaluating predictive uncertainty challenge}, in Machine Learning Challenges,
  Springer, 2006.

\bibitem{Radosavljevic2013}
{\sc V.~Radosavljevic, K.~Ristovski, and Z.~Obradovic}, {\em Gaussian
  conditional random fields for modeling patient's response in acute
  inflammation treatment}, in ICML 2013 workshop on M. L. for Sys. Iden., 2013.

\bibitem{Radosavljevic2010}
{\sc V.~Radosavljevic, S.~Vucetic, and Z.~Obradovic}, {\em Continuous
  conditional random fields for regression in remote sensing}, in ECAI, 2010.

\bibitem{Radosavljevic2014}
\leavevmode\vrule height 2pt depth -1.6pt width 23pt, {\em Neural gaussian
  conditional random fields}, in ECML, 2014, pp.~614--629.

\bibitem{Rasmussen2006}
{\sc C.E. Rasmussen and C.K.I. Williams}, {\em Gaussian Processes for Machine
  Learning}, MIT Press, 2006.

\bibitem{ristovski2013}
{\sc K.~Ristovski, V.~Radosavljevic, S.~Vucetic, and Z.~Obradovic}, {\em
  Continuous conditional random fields for efficient regression in large fully
  connected graphs}, in AAAI, 2013.

\bibitem{Stiglic2014}
{\sc G.~Stiglic, F.~Wang, A.~Davey, and Z.~Obradovic}, {\em Readmission
  classification using stacked regularized logistic regression models}, in
  AMIA, 2014.

\bibitem{Stojanovic2015}
{\sc J.~Stojanovic, M.~Jovanovic, Dj. Gligorijevic, and Z.~Obradovic}, {\em
  Semi-supervised learning for structured regression on partially observed
  attributed graphs}, in SDM, 2015.

\bibitem{Uversky2013}
{\sc A.~Uversky, D.~Ramljak, V.~Radosavljevic, K.~Ristovski, and Z.~Obradovic},
  {\em Which links should i use?: a variogram-based selection of relationship
  measures for prediction of node attributes in temporal multigraphs.}, in
  ASONAM, 2013.

\bibitem{Uversky2014}
\leavevmode\vrule height 2pt depth -1.6pt width 23pt, {\em Panning for gold:
  using variograms to select useful connections in a temporal multigraph
  setting.}, Soc. Netw. Analy. and Mining,  (2014).

\bibitem{Zhou2013}
{\sc J.~Zhou, J.~Sun, Y.~Liu, J.~Hu, and J.~Ye}, {\em Patient risk prediction
  model via top-k stability selection}, in SDM, 2013.

\end{thebibliography}
\end{document}